\documentclass[letterpaper, 10 pt, conference]{ieeeconf}  

\IEEEoverridecommandlockouts                              
                                                          
\newtheorem{theorem}{Theorem}[section]

\newtheorem{rem}{Remark}

\usepackage{xcolor}

\usepackage{algorithm}
\usepackage{algpseudocode}
\usepackage{graphics} 
\usepackage{subcaption}

\usepackage{epsfig} 
\usepackage{mathptmx} 
\usepackage{times} 
\usepackage{amsmath} 
\usepackage{amssymb}  
\usepackage[mathscr]{eucal}
\usepackage{hyperref}
\DeclareMathAlphabet{\mathcal}{OMS}{cmsy}{m}{n}

\DeclareMathOperator*{\argmin}{arg\,min}
\title{\LARGE \bf
Safe Real-Time Optimization using Multi-Fidelity Gaussian Processes
}

\author{Panagiotis Petsagkourakis$^{1}$, Benoit Chachuat$^{2}$ and Ehecatl Antonio del Rio-Chanona$^{2}$
\thanks{* This project has also received funding from the EPSRC project EP/R032807/1.}
\thanks{$^{1}$Panagiotis Petsagkourakis is with Centre for Process Systems Engineering, University College London, London, United Kingdom
        {\tt\small p.petsagkourakis@ucl.ac.uk}}%
\thanks{$^{2}$Benoit Chachuat and Ehecatl Antonio del Rio-Chanona are with Centre for Process Systems Engineering, Imperial College London, London, United Kingdom 
        {\tt\small  a.del-rio-chanona@imperial.ac.uk, b.chachuat@imperial.ac.uk}}%
}

\begin{document}

\maketitle
\thispagestyle{empty}
\pagestyle{empty}

\begin{abstract}
This paper proposes a new class of real-time optimization schemes to overcome system-model mismatch of uncertain processes. This work's novelty lies on integrating derivative-free optimization schemes and multi-fidelity Gaussian processes within a Bayesian optimization framework. The proposed scheme uses two Gaussian processes for the stochastic system, one emulates the (known) process model, and another, the true system though measurements. In this way, low fidelity samples can be obtained via a model, while high fidelity samples are obtained through measurements of the system. This framework captures the system's behavior in a non-parametric fashion, while driving exploration through acquisition functions. The benefit of using a Gaussian process to represent the system is the ability to perform uncertainty quantification in real-time and allow for chance constraints to be satisfied with high confidence. This results in a practical approach that is illustrated in numerical case studies, including a semi-batch photobioreactor optimization problem.
\end{abstract}

\section{INTRODUCTION}

The advantages of real-time optimization (RTO) in industrial processes are well known\cite{Darby2011, Camara2016}, however, currently, most implementations rely on heuristics and trial and error to reach optimality\cite{Camara2016}. The fact that complex phenomena cannot be modelled exactly is the biggest challenge in real-time optimization, in this context we refer to the 'plant' as the 'true' system, and therefore refer to 'plant-model mismatch' as the error of our model with respect to the real system (the plant). Real-time optimization relies on the development of models that are utilized to conduct optimization \cite{Marlin1997}, these (imperfect) models are usually updated in real-time when measurements are available. With the updated models, the optimization is repeated, and an iterative procedure between model refinement and optimization ensues.
This two-step procedure in RTO schemes is called the model-adaptation strategy.
Despite being the most widely used method in industry, this model-adaptation strategy does not converge to the systems's optimal operating conditions. 
This has led research into different methods in RTO \cite{pr5010003,Chachuat2009,Marchetti2009,tsay2020identification}.
One approach that has shown particular promise in the literature is that of modifier adaptation \cite{Marchetti2009}. Modifier adaptation relies on adding linear functions to the optimization model's cost and constraint functions while keeping a nominal process model. 
This approach has shown to be very efficient in some scenarios, nevertheless, function gradients from process measurements are needed, which are difficult to estimate in practice. 
Variants of modifier adaptation to resolve this issue include recursive update schemes \cite{Gao2005,Marchetti2010}, directional derivatives \cite{Costello2016}, transient process measurements \cite{Speakman2020}. 

The use of derivative-free approaches can lift the issues with the estimation of gradients.
The use of quadratic surrogates is proposed in \cite{Gao2016} fitted on available plant data.
Similarly, data-driven models are investigated in \cite{Singhal2016} based on quadratic surrogates as modifiers for the predicted cost and constraint functions.
The natural extension of such surrogates is Gaussian processes (GPs) that were first used in RTO \cite{Ferreira2018}.
Subsequently, GPs were employed in RTO using Bayesian optimization ideas \cite{ CHANONA2021107249}.
Recently, the convergence certificates for such schemes were verified \cite{Shukla2020}. 
The idea of correcting the mismatch of a knowledge-driven model with a data-driven model is akin to hybrid semi-parametric modelling \cite{Thompson1994}, specifically a parallel hybrid model structure. The consideration of non-parametric models, whereby the nature and number of parameter areas not determined by a priori knowledge but tailored to the data at hand, makes perfect sense to capture the structural plant-model mismatch in RTO applications. In principle, this approach is even amenable to a completely model-free RTO scheme by simply discarding the first-principles model component. 

The utilization of non-parametric models such as GPs, where the number of parameters is not determined a priori but rely on the available data, is the most natural approach to capture the structural plant-model mismatch of a system.
In fact, a completely model-free RTO scheme could be used by discarding the mechanistic model component; however, relying on knowing information about the system in the form of a model increases performance \cite{CHANONA2021107249, PETSAGKOURAKIS2020106649, petsagkourakis2020chance}.

\subsection{Multi-fidelity modelling via Gaussian processes}

A process model with low predictive capabilities combined with the real system's samples (or a high fidelity model) has gained a lot of attention in multi-fidelity modelling. Notice that most industrially relevant models are not analytical. They are often complex black-box simulators and legacy code, e.g. systems of ordinary or partial differential equations or a set of `if/else if' rules.
We can evaluate the models, but their derivatives with respect to the optimization variables cannot be trivially found. For this reason, a multi-fidelity Gaussian process is proposed to account for plant-model mismatch and the low fidelity black-box models.  
Popular techniques to perform such a multi-fidelity approach include classical Auto-Regressive (AR) schemes \cite{LIU2018102}.
The linear autoregressive information scheme introduced by Kennedy and O'Hagan \cite{doi:10.1111/1467-9868.00294} has been widely used due to its ease in implementation.
Recently, non-linear autoregressive techniques \cite{Perdikaris2017}, and deep GPs \cite{cutajar2019deep} have been proposed to improve multi-fidelity predictive capabilities. 
Ideas of Multi-fidelity GP have also been used for Bayesian optimization \cite{ pmlr-v115-wu20a}.

In the high model mismatch context, three main frameworks have been proposed in model predictive control (MPC) \cite{Hewing2020}; Learning the dynamics of the system, learning the controller, and safe learning-based MPC. Most techniques cannot easily ensure that safety constraints are met, particularly during learning iterations.
In the context of only parametric uncertainty, it is common to use Polynomial chaos expansions and reformulate the chance constraints using a deterministic surrogate \cite{BRADFORD2019434, Fagiano2012}.
Under the presence of structural mismatch, it is common to use non-parametric models like GPs and incorporates the chance constraint satisfaction in a similar manner \cite{Bradford2020}. 
Similarly, safe Bayesian optimization techniques have been proposed in \cite{Berkenkamp2016}, where the GP's variance is employed to help the safe exploration.
Such approaches are sensible. However, there is no free lunch, in safe exploration; the GP's confidence intervals may not include the true system, especially under the presence of limited data, resulting in the violation of constraints.

In this paper, we propose the use of multi-fidelity GPs to combine available models (low-fidelity) and noisy measurements from the true plant (high-fidelity).
Additionally, chance constraints are reformulated using the Chebyshev-Cantelli theorem \cite{marshall11} to guarantee constraint satisfaction with high confidence.
A trust-region is implemented to restrict the design space and avoid aggressive extrapolations on the predictions and the risky implementation of such an approach.
The trust-regions have an added benefit in terms of online computational costs during the multi-fidelity GP training, as there is no need for many samples from the low fidelity model to be computed due to a restricted design space. 

The rest of the paper provides background on real-time optimization and GPs in Section~\ref{sec:Problem Satement}, and presents convenient background in Section~\ref{sec:background}. In Section \ref{sec:Proposed}, the proposed framework is presented. Then, this algorithm is illustrated with practical case studies in Section~\ref{sec:case}, before drawing final remarks in Section~\ref{sec:conc}.


\section{PROBLEM STATEMENT}\label{sec:Problem Satement}

The optimization of a given plant subject to operational or safety constraints under a stochastic environment can be formulated as:
\begin{align}
\min_{{\bf u}\in\mathcal{U}}~ & G^{\rm p}_0\left({\bf u}\right) := \mathbb{E}\left(g_0\left({\bf u},{\bf y}^{\rm p}({\bf u})\right)\right)\label{eq:plant_problem}\\
\text{s.t.}~ & G_c^{\rm p}\left({\bf u}\right) := \mathbb{P}(g_i\left({\bf u},{\bf y}^{\rm p}({\bf u})\right)\leq 0, \quad i=1\ldots n_g)\geq 1-\alpha \nonumber
\end{align}
where ${\bf u}\in\mathbb{R}^{n_{u}}$ and ${\bf y}^{\rm p}\in\mathbb{R}^{n_{y}}$ are vectors of the plant input (design) and output (measured) variables, respectively. {Notice that the objective is minimized in expectation and the constraints are satisfied with a  pre-defined probability, as the plant is assumed to be affected by Gaussian disturbance and noise.}
Additionally, $g_i:\mathbb{R}^{n_{u}}\times\mathbb{R}^{n_{y}}\rightarrow\mathbb{R}$, $i=0...,n_{g}$, denote the cost and inequality constraint functions; and $\mathcal{U}\subseteq\mathbb{R}^{n_u}$ is the control domain, e.g. lower and upper bounds on the input variables, ${\bf u}^{\rm L}\leq{\bf u}\leq{\bf u}^{\rm U}$. 
The superscript $\left(\cdot\right)^{\rm p}$ is used to indicate plant-related quantities.
Additionally, here $\left(1- \alpha\right)\in \left(0,1\right)$ is a user-defined parameter representing probability of constraint satisfaction with values close to 1.

The challenge in RTO is that the exact mapping ${\bf y}^{\rm p}(\cdot)$ is unknown, and the output ${\bf y}^{\rm p}({\bf u})$ can only be measured for a given input variable ${\bf u}$, under the presence of noise and disturbance. It is common to have an available model (low fidelity) of the plant's input-output behavior ${\bf y}({\bf u},\cdot)$. Nevertheless, the available model may not have a closed-form expression, as they may be complex black-box simulators or a legacy code. 

In the presence of plant-model mismatch and process disturbances, the optimal solution of  Problem  (\ref{eq:plant_problem})  could be significantly different from the optimization that utilizes only the model. 
\section{Background}\label{sec:background}
\subsection{Gaussian Process}\label{sec:subsection_GP}
In this section, Gaussian processes (GPs) are introduced, as one of the key components of the methodology that follows.
The Gaussian process generalizes the multivariate Gaussian distribution to a distribution over an infinite-dimensional vector of functions, such that every finite sample of function values are jointly Gaussian distributed.
Due to the Bayesian nature, GPs can consider both epistemic  (limited data) and aleatoric (stochasticity of the true model) uncertainty.
A GP is fully specified with the prior mean function $\mu(\cdot)$ and the positive semi-definite kernel function $k(\cdot,\cdot)$.
The GP regression aims to model an unknown set of functions $\mathbf{g}:\mathbb{R}^{n_u}\rightarrow\mathbb{R}^{n_y}$ given some noisy observations $\textbf{y}=g(\textbf{u})+\epsilon$. 
For $\mathbf{g}=\left[g_1,…,g_{n_y}\right]$, this could be expressed as:
\begin{equation}
    g_i\sim \mathcal{G}\mathcal{P}(\mu_i(\cdot),k_i(\cdot,\cdot)),
\end{equation}
where the prior mean $\mu_i$ function provides knowledge of the mean of a test point prior to observing data, the kernel function $k_i$ expresses the covariance between points. 
Let a number of data points be available, the posterior distribution at a test point $\textbf{u}^*$ is then found by the conditional distribution given the available $N$ noisy data for each output $i$:
$\{\textbf{U},\textbf{y}_i\}$, with $\textbf{U}=\left[\textbf{u}_1,\dots,\textbf{u}_N\right]\in \mathbb{R}^{n_u\times N}$ and $\textbf{y}_i=\left[{y}_{i1},\dots,{y}_{iN}\right]^T\in \mathbb{R}^{N}$.
Then, the posterior mean $m_i({\textbf{u}^*})$  and variance $\Sigma_i({\textbf{u}^*})$ of the test point $\textbf{u}^*$ are
\begin{equation}\label{eq:posterior}
\begin{split}
    m_i({\textbf{u}^*}) &= \textbf{k}_i^* \left[\textbf{K}_i  + \sigma_i^2 I\right]^{-1}(\textbf{y}_i -\mathbf{1}\mu_i({\textbf{u}^*}))+ \mu_i({\textbf{u}^*})\\
    \Sigma_i({\textbf{u}^*}) &= k_i({\textbf{u}^*},{\textbf{u}^*}) - \textbf{k}_i^* \left[\textbf{K}_i  + \sigma_i^2 I \right]^{-1}\textbf{k}_i^{*T},
\end{split}
\end{equation}
where the posterior mean is $m_i({\textbf{u}^*}) = \mathbb{E}(f_i^*|{\textbf{u}^*},\textbf{U},\textbf{y}_i)$, the posterior variance is $\Sigma_i({\textbf{u}^*}) = \mathbb{V}(f_i^*|{\textbf{u}^*}, \textbf{U},\textbf{y}_i)$, $\textbf{k}_i^* = \textbf{k}_i({\textbf{u}^*},\textbf{U})$ is the covariance between training and test cases vector, $\textbf{K}_i = \textbf{k}_i(\textbf{U},\textbf{U})$ is covariance (or Gram) matrix.
Notice that $\Sigma_i$ is the variance of the noise-free prediction, if the noise is taken into consideration then variance should be $\Sigma_i + \sigma_i^2$, and $\sigma_i^2$ is many times treated as a hyperparameter for the GP.
The predicted distribution of  $\textbf{g}({\textbf{u}^*})$ follows a normal distribution:
\begin{equation}
    \textbf{g}({\textbf{u}^*}) \sim \mathcal{N}(\textbf{m}({\textbf{u}^*}),\pmb{\Sigma}({\textbf{u}^*})),
\end{equation}
with $\textbf{m}({\textbf{u}^*}) =\left[m_1,\dots,m_{n_y}\right]$ and $\pmb{\Sigma}({\textbf{u}^*}) = \text{diag}(\left[\Sigma_1({\textbf{u}^*}),\dots,\Sigma_{n_y}({\textbf{u}^*})\right])$.
Notice, that multi-output formulation of the GP could be use instead to compute the non-diagonal terms of the covariance \cite{alvarez2012kernels}.
Common choices for the kernel $k_i(\cdot,\cdot)$ is the squared-exponential (SE) covariance function
and Mat{\'{e}}rn class of covariance functions, with special cases the  Mat{\'{e}}rn $3/2$ and  Mat{\'{e}}rn $5/2$ \cite{Rasmussen2006}
with the hyperparamters and the noise level  being estimated using maximum likelihood estimation.

\subsection{Multi-fidelity Gaussian process}\label{sec:Multi-fidelity GP}

The GP regression framework can be systematically extended to formulate probabilistic models that enable different fidelity information sources (see \cite{Perdikaris2017, LIU2018102}), i.e. a low-fidelity model for which many data points can be generated and a high-fidelity model/system for which only a few data-points can be obtained.
Generally, numerous levels of fidelity models can be included, in this work, 2 levels of fidelity are considered, i.e. the low fidelity (known model) and the high fidelity (plant measurements).

Suppose that we have data points for the low $y^{\rm m}$ and high fidelity $y^{\rm p}$ system at locations $\textbf{u}^{\rm m}\in \mathcal{D}^{\rm m}$ and $\textbf{u}^{\rm p}\in \mathcal{D}^{\rm p}$, respectively.
The collected data can be organized as pairs by increasing fidelity as $\mathcal{D}_t = \{\textbf{u}^{\rm t}\}$, $Y^{\rm t} =\{{y}^{\rm t}\}$,
$t = \{\rm m, p\}$. The prediction model can be written as an auto-regressive scheme: 
\begin{equation}\label{eq:multifidelity}
    g^{\rm p}(\textbf{u}) = \epsilon g^{\rm m}(\textbf{u}) + \delta(\textbf{u})
\end{equation}
where $g^{\rm m}$ and $g^{\rm p}$ are GPs modelling the data at low and high fidelity respectively.
Additionally, $\epsilon$ is a scaling constant that can be estimated and $\delta(\cdot)$  is a GP with mean $\mu_\delta(\cdot)$ and covariance $k_\delta(\cdot,\cdot)$ functions , i.e. $ \delta \sim \mathcal{G}\mathcal{P}(\mu_\delta(\cdot), k_\delta(\cdot,\cdot))$. 

A numerically efficient inference and hyper-parameter learning scheme can be employed by assuming that the experimental design sets ${\mathcal{D}_m}$ and $ \mathcal{D}_p$ have a nested structure, i.e. $\mathcal{D}_p\subseteq \mathcal{D}_m$ \cite{Gratiet2014}.
This assumption means that the lower fidelity's training inputs should include the data of the higher fidelity level.
The inference problem is now decoupled into 2 standard GP regression problems, resulting in the multi-fidelity posterior distribution with predictive mean and variance at each level for a given $\textbf{u}^*$ 
given by
\begin{equation}\label{eq:posterior_multifidelity}
\begin{split}
    m_{g^{\rm p}}({\textbf{u}^*}) &=
    \textbf{k}_\delta^* \left[\textbf{K}_\delta  + \sigma_\delta^2I \right]^{-1}(\textbf{y}^{\rm p} 
    -\mathbf{1}(\epsilon m_{g^{\rm m}}({\textbf{u}^*}) \\&+ \mu_\delta({\textbf{u}^*}) ))+ \epsilon m_{g^{\rm m}}({\textbf{u}^*}) + \mu_\delta({\textbf{u}^*})\\
    \Sigma^{\rm p}({\textbf{u}^*}) &=
    \epsilon^2 \Sigma^{\rm m}({\textbf{u}^*}) + k_\delta({\textbf{u}^*},{\textbf{u}^*}) -\textbf{k}_\delta^* \left[\textbf{K}_\delta  + \sigma_\delta^2 I \right]^{-1}\textbf{k}_\delta^{*T}.
\end{split}
\end{equation}
The training of the lower fidelity model includes the training data points of the plant (higher fidelity)  $\textbf{u}^{\rm p^*}$, as the result the posterior prediction of the low fidelity at $\textbf{u}^{\rm p^*}$ is by construction a deterministic quantity, the model training points are assumed to be noiseless. Hence, due to the nested training sets (i.e.$\mathcal{D}_p\subseteq \mathcal{D}_m$), the training of $g^{\rm p}$ given the available data reduces to a straightforward maximum-likelihood estimation problem.
\subsection{Chance Constraints}
Given the problem definition (\ref{eq:plant_problem}), we wish to satisfy the constraints with a given probability.
These chance constraints can be reformulated using the multi-fidelity GPs.
To satisfy the chance constraints, a distributionally robust reformulation is employed via the Chebyshev-Cantelli theorem \cite{marshall11}, (Theorem \ref{thm:Chebyshev}) using the mean and variance of the constraints.
\begin{theorem}\label{thm:Chebyshev}(\cite{marshall11})
Consider a chance constraint of the form.
\begin{equation}
    \mathbb{P}(q\leq 0)\geq 1-\gamma, ~~~~\gamma\in (0,1)
\end{equation}
where $q\in \mathbb{R}^{n_q}$ is some random variable. Let $\mathcal{Q}$ be a family of distributions  with mean $\mu_q$ and variance $\Sigma_q$. Then for any $\epsilon\in(0,1)$, the distributionally robust probabilistic constraint 
\begin{equation}
    \inf_{q\sim \mathcal{Q}}\mathbb{P}(q\leq 0)\geq 1-\gamma
\end{equation}
is equivalent to the following constraint:
\begin{equation}
    \mu_q + r\sqrt{\Sigma_q}\leq 0,
\end{equation}
with $r=\sqrt{\dfrac{1-\gamma}{\gamma}}$.
\end{theorem}
This theorem provides a conservative estimate for the constraints and it holds for the whole family distribution $\mathcal{Q}$.
\begin{rem}
The parameter $r$ is often used as a design parameter to reduce the inherent conservatism; however, this makes any guarantees void.  
\end{rem}
This reformulation highly relies on the accuracy of the mean and variance computed via a GP, hence the initial small amount of data points may significantly affect the result.
Notice that even though the result in Theorem \ref{thm:Chebyshev} is conservative for all the family of distributions with mean $\mu_q$ and $\Sigma_q$, the wrong estimation of mean and variance may lead to infeasible designs.

To overcome this limitation, trust-regions are employed, where the design space of the optimization variables $\textbf{u}$ is further restricted.

\section{Proposed Framework}\label{sec:Proposed}
\subsection{Trust Regions}\label{sec:TR}

The use of GPs has gained a lot of attention due to their ability to provide a close form expression for both the mean and variance; 
however, their use far from collected data (global use of the GP) may be naive.
To avoid this, trust regions are utilized to update the GPs for the model-mismatch locally and not globally. This is particularly important in safety-critical scenarios, such as in real-time optimization and many engineering applications.

To motivate the reader on the use of trust regions, we introduce the assumptions that are usually utilized for GPs as global surrogates; then, we show that even for a simple example, such an approach may not be reliable for a global prediction.

It is common to achieve bounds for the unknown function, {assuming that the unknown function ${g}_i$ has bounded norm in reproducing kernel Hilbert space (RKHS) \cite{Sui2015}. This assumption\cite{Srinivas2010} can lead to reliable confidence intervals.}



{ Due to the generic nature of the systems, the assumption and confidence interval from \cite{Srinivas2010} may not  globally hold and invalidate the use of GPs globally. To further indicate the importance of the trust-region, a motivating example is presented next.} 

\subsubsection{Importance of Trust Regions: A motivating example }

\begin{figure}[h]
    \centering
    \includegraphics[scale=0.5]{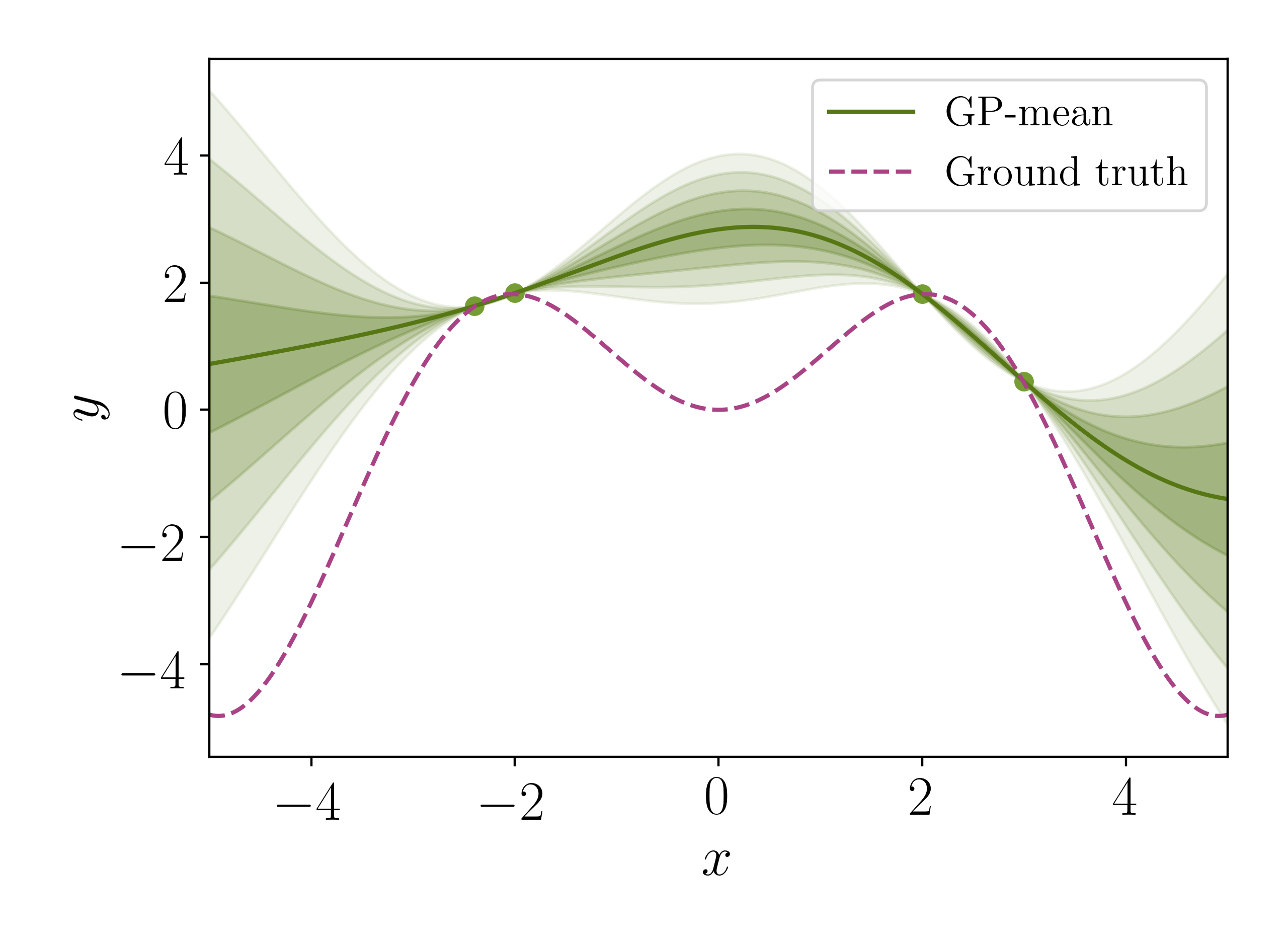}
    \caption{The ground truth (red dashed line) is the function that the GP approximates given the available data. The GP predictions for the mean ($\mu$) is the green line with variance $\Sigma$, and the shaded areas represents the area between $m\pm\sqrt{\Sigma}, m\pm\sqrt{\Sigma}, \mu\pm3\sqrt{\Sigma}, m\pm4\sqrt{\Sigma}$ and  $m\pm10\sqrt{\Sigma}$.}
    \label{fig:simple_GP_danger}
\end{figure}

%
Let a simple function $f(x) = x \sin(x)$ be assumed as the `true' system, with $x\in \mathbb{R}$ and the collected data being corrupted by a Gaussian noise $\epsilon_f$ with standard deviation 0.01, $y=f(x) + \epsilon_f$. 
A GP is used to approximate the data; a squared-exponential (SE) covariance function is used and its hyperparameters are fitted using maximum likelihood estimation (with multi-starts to avoid highw laying local optima).
The mean $m$ and variance $\Sigma$ of the GP for each $x$ are illustrated in Fig~\ref{fig:simple_GP_danger}. 

The shaded areas show the $m+r \sqrt{\Sigma}$ for $r=1, 2, 3, 4$ and $10$.
If we were to use Theorem~\ref{thm:Chebyshev}, we would expect the `true' model (red dashed line) to lie within these shaded areas.
The Chebyshev-Cantelli theorem states that for $r=9.94$, $x$ should lie within the shaded area with a probability of $99\%$. However, due to the shape of the  `true' model and the collected data, the predictions are poor.
This example motivates the restriction of the proposed optimization problem.
Although this is a well know fact for practitioners, it not generally taken into account, and it is a particular pitfall of any approach that wishes to satisfy constraints with high probability.
%
%
%
\begin{rem}[Trust-regions for multi-fidelity GPs]
The advantages of the trust-region are not limited to the ability of the GP to accurately represent the system and the reformulation of the chance constraints. 
The construction of the GP for the low fidelity model requires the acquisition of data from a simulator, which is generally expensive in large and complex processes.
Without trust regions, the whole design space must be explored, which means that either i) the number of data points is not enough to provide a good approximation, and the posterior variance is high, significantly affecting both the acquisition function and the reformulated chance constraints or ii) a large amount of data points is drawn, incurring in high computational costs for the sampling, training of the hyperparameters and prediction, which is intractable in many instances. Variational methods \cite{pmlr-v5-titsias09a} could be used to mitigate this issue; however, the problem remains as the number of design variables increases. 

This issue is accommodated with the use of the trust-region, as only data in the trust-region's neighbourhood in needed.

\end{rem}

\subsection{Proposed framework}

Herein, we introduce the concept of trust-regions from the field of derivative-free optimization together with acquisition functions from Bayesian optimization and chance constraints for safe exploration.
Let $g^{\rm p}_i$ be the plant's cost ($i=0$) and inequality constraints ($i=1,\dots, n_g$), $g^{\rm m}_i$ be the model's cost and inequality constraints, and $m_{g_i^{\rm p}}, {\Sigma_i^{\rm p}}, m_{g_i^{\rm m}}, {\Sigma_i^{\rm m}}$ their corresponding GP's posterior mean and variance.
Then, the modified optimization problem that is solved at each RTO iteration becomes:
\begin{equation}
\begin{split}\label{eq:modified_problem_GP+TR}
{\bf d}^{k+1} \in \argmin_{\bf d}&~  \mathcal{A}[m_{g_0^{\rm p}}, {\Sigma_0^{\rm p}}]({\bf u}^k+{\bf d}) \\
\text{s.t.}~ & [m_{g_i^{\rm p}}+r\sqrt{\Sigma_i^{\rm p}}]({\bf u}^k+{\bf d}) \leq 0,\\
& \|{\bf d}\|\leq\Delta^{k}, \quad {\bf u}^k+{\bf d}\in\mathcal{U} \\
&r = \sqrt{\dfrac{1-\alpha}{\alpha}},
\quad i=1\ldots n_g 
\end{split}
\end{equation}
where $\Delta^k{>} 0$ is the trust-region radius for the predicted step ${\bf d}^{k+1}\in\mathbb{R}^{n_u}$; and $\mathcal{A}$ is an acquisition function for the cost associated with the its posterior mean and variance ($m_{g_i^{\rm p}}, {\Sigma_i^{\rm p}}$){, leading to, for example,} the Lower Confidence Bound (LCB) or Expected Improvement (EI) function\cite{Jones1998, CHANONA2021107249} .

The solution of Problem~\eqref{eq:modified_problem_GP+TR} coincides with a constrained Bayesian optimization within a trust-region.
The trust-region is adapted when a new sample becomes available according to an adaptation mechanism (see next section \ref{sec:adaptation} and Algorithm~\ref{alg:MA-GP-TR}).

\subsubsection{Adaptation Mechanisms}\label{sec:adaptation}
As stated in section \ref{sec:TR}, the trust region's role is to restrict the step size of the next design, so that cost and constraints are accurate predictions of the true system.
The trust region is updated according to classical update rules in trust-region algorithms \cite{Conn2009}, which relies on the ratio of the actual cost reduction versus the predicted cost reduction:
\begin{align}
\rho^{k+1} := \frac{g_0^{\rm p}\left({\bf u}^{k}\right)-g_0^{\rm p}\left({\bf u}^{k}+{\bf d}^{k+1}\right)}{m_{g_0^{\rm p}}({\bf u}^{k})-m_{g_0^{\rm p}}({\bf u}^{k}+{\bf d}^{k+1})} \label{eq:merit_function}
\end{align}
The trust-region radius $\Delta^{k+1}$ is reduced whenever the accuracy ratio $\rho^{k+1}$ is too low. 
Conversely, $\Delta^{k+1}$ is increased if the solution of \eqref{eq:modified_problem_GP+TR} takes a full step and the prediction of the plant cost is good around this point.
In a different scenario, the trust-region radius stays unchanged.
Similarly, the new design point ${\bf u}^{k}+{\bf d}^{k+1}$ is accepted if the step ${\bf d}^{k+1}$ produces an accuracy ratio $\rho^{k+1}$ which is sufficiently large.
Otherwise, the operating point remains unchanged. In practical situations this would trigger a back-tracking {from ${\bf u}^{k}+{\bf d}^{k+1}$ to ${\bf u}^{k}$} during the RTO cycle. 
Any infeasibility trigger a rejection of the current step ${\bf d}^{k+1}$ as well {and, result in backtracking to point ${\bf u}^{k}$}.
Notice that our approach considers a chance constraint formulation which means that the constraint violation does not occur often.

\subsubsection{Algorithm}

\begin{algorithm}[H]
\caption{RTO using multifidelity GPs}\label{alg:MA-GP-TR}
{\bf Input:} initial data sets $({\bf U}^0,{\bf g}_i^{{\rm p} 0})$, $i=0\ldots n_g$; initial operating point $\mathbf{u}^0\in\mathcal{U}$; initial and maximal trust-region radii $0<\Delta^0<\Delta_{\rm max}$; trust-region parameters $0<\eta_1<\eta_2<1$, $0 < \gamma_{\rm red} < 1 < \gamma_{\rm inc}$, ${\alpha}>0$, the number of $N_{m}$ extra simulations from the black-box model.\\
I) Generate the nested set (i.e.$\mathcal{D}_p\subseteq \mathcal{D}_m$), with $N_{m}$ additional simulations in the neighborhood of the trust-region.\\
II) Construct the GP posterior for the low and high fidelity GP,
 $\rhd$\ $m_{g_i^{\rm p}}, {\Sigma_i^{\rm p}}, m_{g_i^{\rm m}}, {\Sigma_i^{\rm m}}$

{\bf Repeat: for $k=0,1,\ldots$} 
\begin{enumerate}

\item Solve modified optimization problem (Problem~\ref{eq:modified_problem_GP+TR})
 $\rhd$\ $\mathbf{d}^{k+1}$

\item Get process cost and constraint measurements
 $\rhd$\ $G^{\rm p}_i({\bf u}^{k}+{\bf d}^{k+1}), i=0\ldots n_g$

\item Check infeasibility\\
If {either} Problem~\eqref{eq:modified_problem_GP+TR} is infeasible, or {if }$g_{i}^{\rm p}({\bf u}^{k}+{\bf d}^{k+1})>0$ for any $i>0$:\\
${\quad} \Delta^{k+1}\ \leftarrow\ [\gamma_{\rm red},1] \Delta^{k}$, \quad ${\bf u}^{k+1}\ \leftarrow\ {\bf u}^k\ \text{(reject)}$, \quad and go to Step 7

\item Compute merit function  (Equation~\ref{eq:merit_function})
 $\rhd$\ $\rho^{k+1}$

\item Update trust region\vspace{-.5em}
\begin{align*}
& \text{If $\rho^{k+1} > \eta_2\ \wedge\ \|{\bf d}^{k+1}\|=\Delta^{k}$:} \\ & \quad\Delta^{k+1}\ \leftarrow\ \gamma_{\rm inc} \Delta^{k}, \quad {\bf u}^{k+1}\ \leftarrow\ {\bf u}^k + {\bf d}^{k+1}\ \text{(accept)} \\
& \text{Else If $\rho^{k+1} < \eta_1$:}
\\ &\quad \Delta^{k+1}\ \leftarrow\ \gamma_{\rm red} \Delta^{k},\quad {\bf u}^{k+1}\ \leftarrow\ {\bf u}^k\ \text{(reject)}\\
& \text{Else:}\quad \Delta^{k+1}\ \leftarrow\  \Delta^{k},  \quad \quad{\bf u}^{k+1}\ \leftarrow\ {\bf u}^k + {\bf d}^{k+1}\ \text{(accept)}
\end{align*}

\item Update data set for the plant
 $\rhd$\ $\mathcal{D}_p = \{\mathcal{D}_p\cup{\bf u}^{k+1}\},  \textbf{Y}_i^{{\rm p}}= \{ \textbf{Y}_i^{{\rm p}}\cup{ g}_i^{{\rm p}k+1}\} i=0\ldots n_g$
\item Generate the nested nested training set (i.e.$\mathcal{D}_p\subseteq \mathcal{D}_m$), with $N_{m}$ additional simulations in the neighborhood of the trust-region.
\item Update low and high fidelity GP,
 $\rhd$\ $m_{g_i^{\rm p}}, {\Sigma_i^{\rm p}}, m_{g_i^{\rm m}}, {\Sigma_i^{\rm m}}$
\end{enumerate}
\end{algorithm}

In {\bf steps I)} and {\bf II)} the training set from the model and the low-fidelity GPs are constructed. The RTO iterations begin, where the first optimization is solved in {\bf step 1}.
Then the new optimized variables are applied to the true system and the corresponding measurements are sampled in {\bf step 2}.
Then, {\bf steps 3-5} follow the adaptation procedure described in section \ref{sec:adaptation}.
Apart from updating the trust region, both the data sets and the GPs are updated in {\bf Steps 6, 7} and {\bf 8} respectively, this is irrespective of whether the step ${\bf d}^{k+1}$ is accepted or not. 
Specifically, at {\bf step 6}, the data points of the plant are generated, i.e. measurements from the real system are collected.
Then in {\bf step 7} the nested set is constructed, where $N_m$ random simulations are performed using the black-box model in the neighborhood of the trust-region. 
The black-box is also simulated at the new operational point $\textbf{u}^{k+1}$. Hence, the input training set of the model $\mathcal{D}_m$ consists of the $N_m$ as well as the set $\mathcal{D}_p$. {The additional $N_m$ samples are randomly generated inside the trust-region, as it is important for the surrogate to be accurate inside the trust region.}
The computational burden of reconstructing the GPs at each iteration could be eased upon updating the covariance matrix at certain iterations only \cite{Rasmussen2006}.

\section{CASE STUDY: Batch-to-Batch Optimization}\label{sec:case}


This case study investigates the performance of the proposed methodology in a high-dimensional RTO problem.
We consider the batch-to-batch optimization of a photobioreactor for the production of phycocyanin ({\sf P}) by the blue-green cyanobacterium {\em Arthrospira platensis} ({\sf X}) growing on nitrates ({\sf N}).
A dynamic model describing the concentrations $C_{\sf X}\ \rm [g\,L^{-1}]$, $C_{\sf N}\ \rm [mg\,L^{-1}]$ and $C_{\sf P}\ \rm [mg\,L^{-1}]$ in the  photobioreactor is given by \cite{Bradford2020}: 
\begin{align}
\dot{C}_{\sf X} =\ & u_{\sf m} \dfrac{I(t)}{I(t) + k_{\sf s} + I(t)^2/k_{\sf i}} \dfrac{ C_{\sf N}(t)}{C_{\sf N}(t) + K_{\sf N}}C_{\sf X}(t) - u_{\sf d}C_{\sf X}(t)\label{ode:X}\\
\dot{C}_{\sf N} =\ & -Y_{\sf N/X} u_{\sf m} \dfrac{I(t)}{I(t) + k_{\sf s} + I(t)^2/k_{\sf i}} \dfrac{ C_{\sf N}(t)}{C_{\sf N}(t) + K_{\sf N}}C_{\sf X} + F_{\sf N}(t) \label{ode:N}\\
\dot{C}_{\sf P} =\ & k_{\sf m} \dfrac{I(t)}{I(t) + k_{\sf sq} + I(t)^2/k_{\sf iq}}{C}_{\sf X}(t) - k_{\sf d} \dfrac{C_{\sf P}(t)}{C_{\sf N}(t) + K_{\sf Np}} \label{ode:P}
\end{align}
%
%
where the light intensity $I(t)$ $\rm [\mu E\,m^2\,s^{1}]$ and the nitrate inflow rate $F_{\sf N}(t)$ $\rm [mg\,L^{-1}\,h^{-1}]$ are manipulated inputs;
and the values of the model parameters $k_{\sf d}$, $k_{\sf m}$, $k_{\sf s}$, $k_{\sf i}$, $k_{\sf sq}$, $k_{\sf iq}$, $K_{\sf N}$, $K_{\sf Np}$, $u_{\sf d}$, $u_{\sf m}$, $Y_{\sf N/X}$ are the same as those reported by \cite{Bradford2020}.
For simplicity, the mass-balance equations \eqref{ode:X}--\eqref{ode:P} neglect the change in volume due to the nitrate addition kinetic model assumes nutrient-replete growth conditions.

The optimization problem seeks to maximize the end-batch concentration of phycocyanin after 240 hours of operation. Regarding constraints, the phycocyanin-to-cyanobacterial-biomass ratio must be kept under $1.1\ \rm wt\%$ at all times; the nitrate concentration must be kept under $800\ \rm mg\,L^{-1}$ at all times and below $150\ \rm mg\,L^{-1}$ at the end of the batch; and both manipulated inputs are bounded. A mathematical formulation of this (dynamic) optimization problem is as follows:
\begin{align}
\min_{I(t),F_{\sf N}(t)}~~ & \mathbb{E}\left(C_{\sf P}(240)\right)\label{eq:PBR}\\
\text{s.t.}~~ & \text{PBR model \eqref{ode:X}--\eqref{ode:P}}\nonumber\\
& C_{\sf X}(0) = 1,\ C_{\sf N}(0) = 150,\ C_{\sf P}(0) = 0\nonumber\\
& \mathbb{P}\left(C_{\sf P}(t) \leq 0.011 C_{\sf X}(t)\right)\geq 0.9,\ \forall t\nonumber\\
& \mathbb{P}\left(C_{\sf N}(t) \leq 800\right)\geq 0.9,\ \forall t\nonumber\\
& 120 \leq I(t) \leq 400,\ \forall t\nonumber\\
& 0 \leq F_{\sf N}(t) \leq 40,\ \forall t\nonumber
\end{align}
To recast it as a finite-dimensional optimization problem, both control trajectories are discretized using a piecewise-constant parameterization over 6 equidistant stages (of 60 hours each). The batch-to-batch optimization, therefore, comprises a total of 12 degrees of freedom. The state path constraints are also discretized and enforced at the end of each control stage.

The case study assumes that the concentrations $C_{\sf X}$, $C_{\sf N}$ and $C_{\sf P}$ can all be measured during or at the end of the batch as necessary.
Process noise is simulated in this virtual environment by adding white noise with zero mean and standard deviation $\sigma_{C_{\sf X}}= 0.02\ \rm [g\,L^{-1}]$, $\sigma_{C_{\sf N}}= 0.316\ \rm [mg\,L^{-1}]$, and $\sigma_{C_{\sf P}}= 0.001\ \rm [mg\,L^{-1}]$. However, no prior knowledge of this measurement noise is assumed during the construction of the GP surrogates for the cost and constraints.


Next, we implement several strategies to Problem~\eqref{eq:PBR}, including Algorithm~\ref{alg:MA-GP-TR} with the use of chance constraints and satisfaction in expectation. The following dynamic model is used for the latter, which presents a structural mismatch with the plant model \eqref{ode:X}--\eqref{ode:P} regarding the light inhibition:
\begin{align}
\dot{C}_{\sf X} =\ & u_{\sf m} \dfrac{I(t)}{I(t) + k_{\sf s}} \dfrac{ C_{\sf N}(t)}{C_{\sf N}(t) + K_{\sf N}}C_{\sf X}(t) - u_{\sf d}C_{\sf X}(t)\label{ode:Xmod}\\
\dot{C}_{\sf N} =\ & -Y_{\sf N/X} u_{\sf m} \dfrac{I(t)}{I(t) + k_{\sf s}} \dfrac{ C_{\sf N}(t)}{C_{\sf N}(t) + K_{\sf N}}C_{\sf X}(t) + F_{\sf N}(t) \label{ode:Nmod}\\
\dot{C}_{\sf P} =\ & k_{\sf m} \dfrac{I(t)}{I(t) + k_{\sf sq}} \dfrac{c_{\sf N}}{C_{\sf N}(t) + K_{\sf N}}C_{\sf X}(t) - k_{\sf d} \dfrac{C_{\sf P}(t)}{C_{\sf N}(t) + K_{\sf Np}} \label{ode:Pmod}
\end{align}
The Mat\'ern kernel (with parameter $\nu=\frac{3}{2}$) is used for both the model and physical system (low and high fidelity, respectively) to construct the multi-fidelity model.
Specifically, for the model, 30 data points are used to tune the respective GP's hyperparameters. 
Notice that now there is no need to involve discretized equations in the NLP problem, as a GP emulates the ode's solution. 
A multi-start initialization (20 random starting points) is applied to overcome the NLP solver's numerical failures and reduce the likelihood of converging to high laying local optima. 
The initial GPs are trained with 13 feasible data points, and the initial trust region encloses all of these points.

\begin{figure}[H]
    \centering
    \includegraphics[width=.98\linewidth]{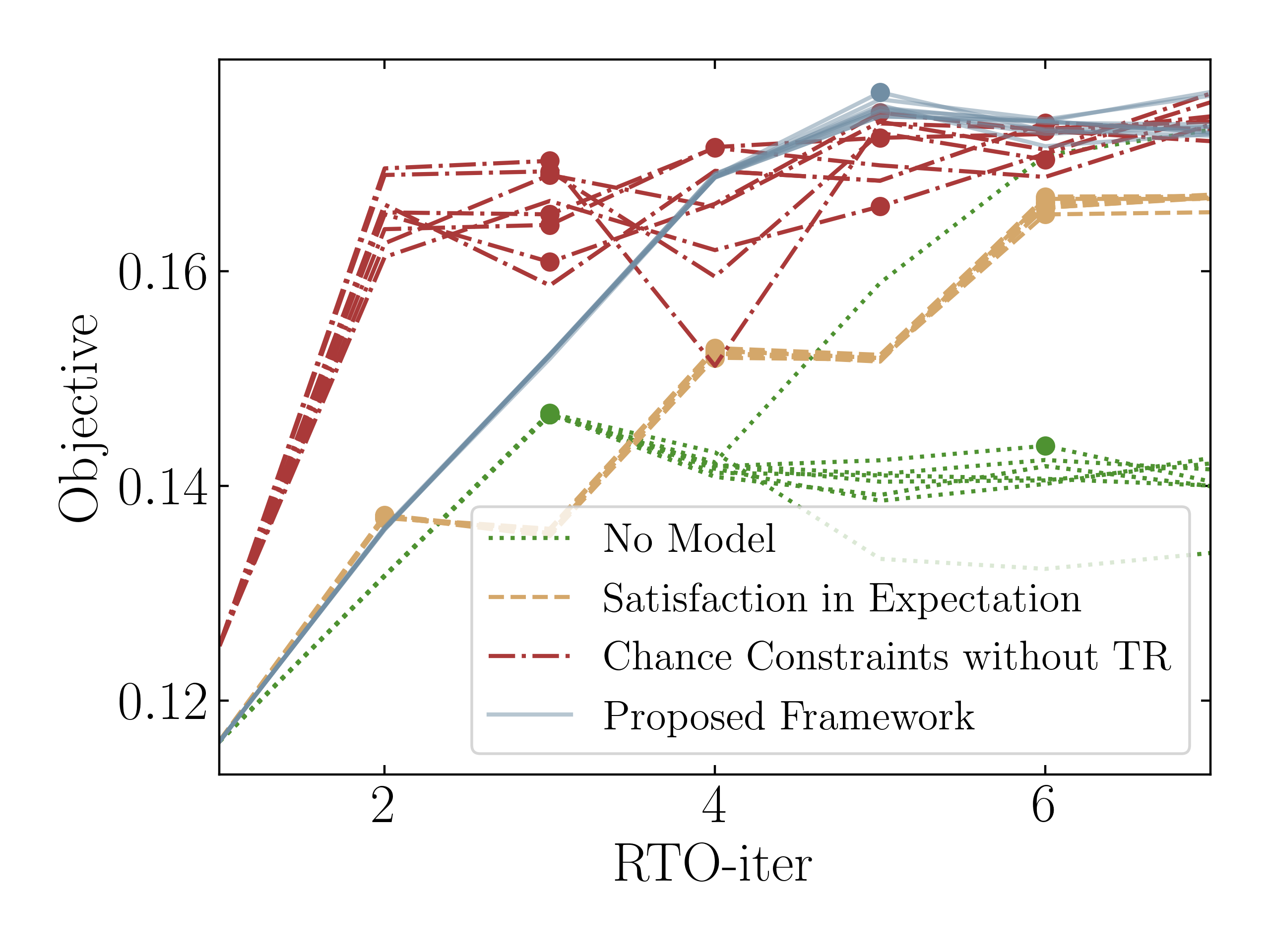}
    \caption{Evolution of process cost with the RTO iterations. Circles represents the violation of a constraint.}
    \label{fig:objective}
\end{figure}

The performance of Algorithm~\ref{alg:MA-GP-TR} using the mulitfidelity GPs and EI acquisition function is compared with three alternative frameworks:\\
{\bf a)} The use of constraint satisfaction in expectation {\it using} trust-regions.\\
{\bf b)} The use of chance constraints {\it without using} trust-regions.\\
{{\bf c)} The use of chance constraints {\it using} trust-regions and {\it without using} a prior model (model-free RTO).}\\
The results for the convergence of the objective are presented in Figure~\ref{fig:objective} for multiple realizations of the process noise.
The circles represent an occurrence of a constraint violation. 
The proposed framework appears to have the best trade-off between convergence and satisfaction of constraints. %
The use of constraint satisfaction in expectation {\bf (a)}, as anticipated, results in increasing constraint violations (compare to the proposed framework); this has a side-effect, it slows down the algorithm as it cannot converge fast to an optimum point as it backtracks often.
On the other hand, the proposed framework converges very fast as the chance constraint and trust-region formulation allow the 'safe' exploration with high probability. 
Scenario {\bf (b)} explores the performance of the algorithm with the removal of the trust-region.
Since the trust region is not restricting the design space, the optimization approaches high values very early in the RTO iterations. However, the GPs are overly confident, and violation of constraints occur often.  
{The last scenario, where an a priori model is ignored {\bf (c)}, violates much less constraints compared to {\bf (a)} and {\bf (b)}, as the optimization problem considers chance constraints and the trust region. However, the absence of prior model in {\bf (c)} affects the convergence of the objective function significantly.}
Notice that the proposed framework managed to violate the least constraints compare to {\bf(a- c)}.
\begin{figure}[H]
    \centering
    \includegraphics[width=.98\linewidth]{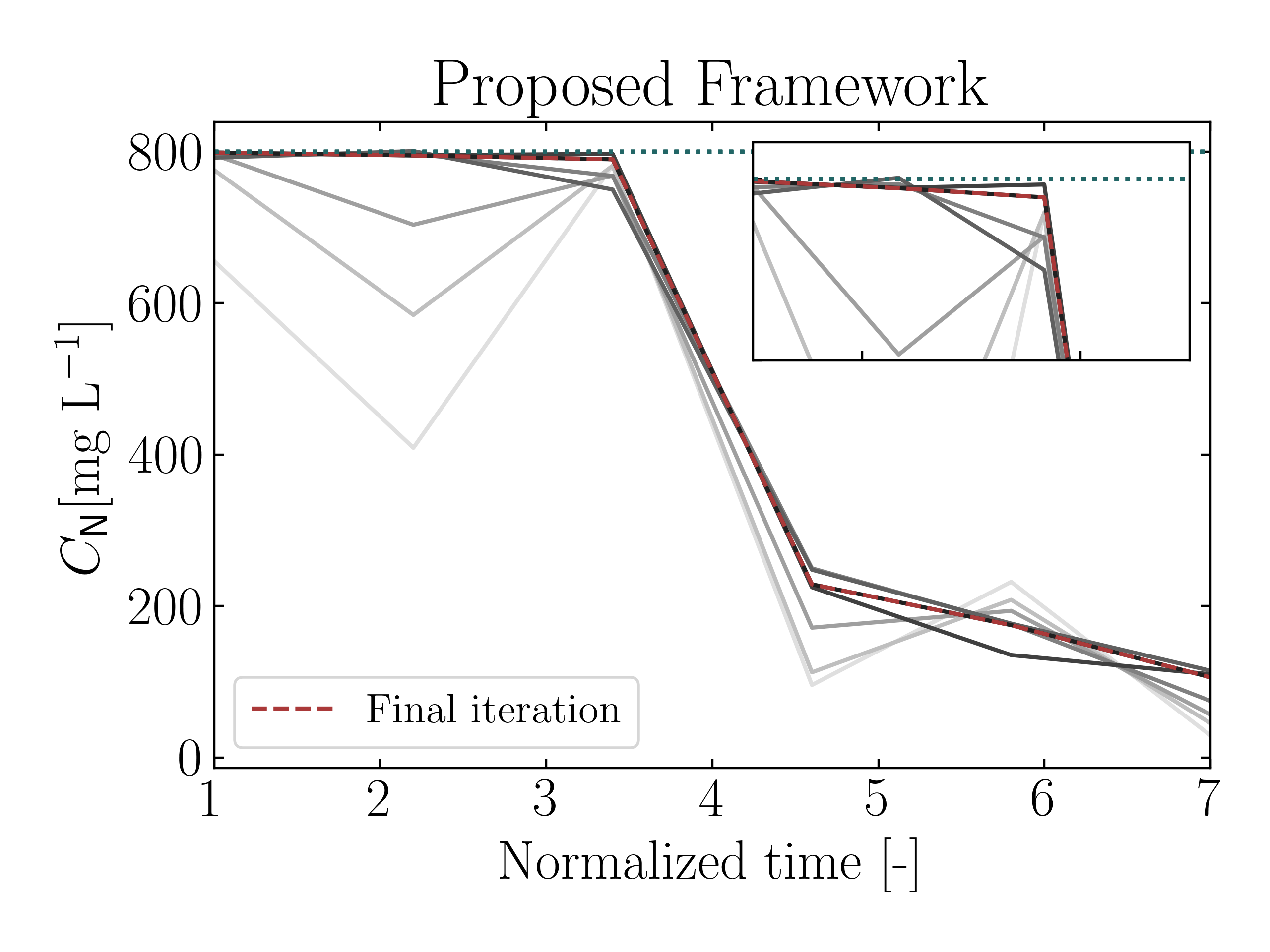}
    \caption{RTO results for the case study (Problem~\ref{eq:PBR}) using our proposed framework. The lines are plotted over the RTO iterations, which are faded out towards earlier iterations }
    \label{fig:constraints}
\end{figure}

\begin{figure}[H]
\begin{subfigure}{0.5\textwidth}
  \centering
  \includegraphics[width=.98\linewidth]{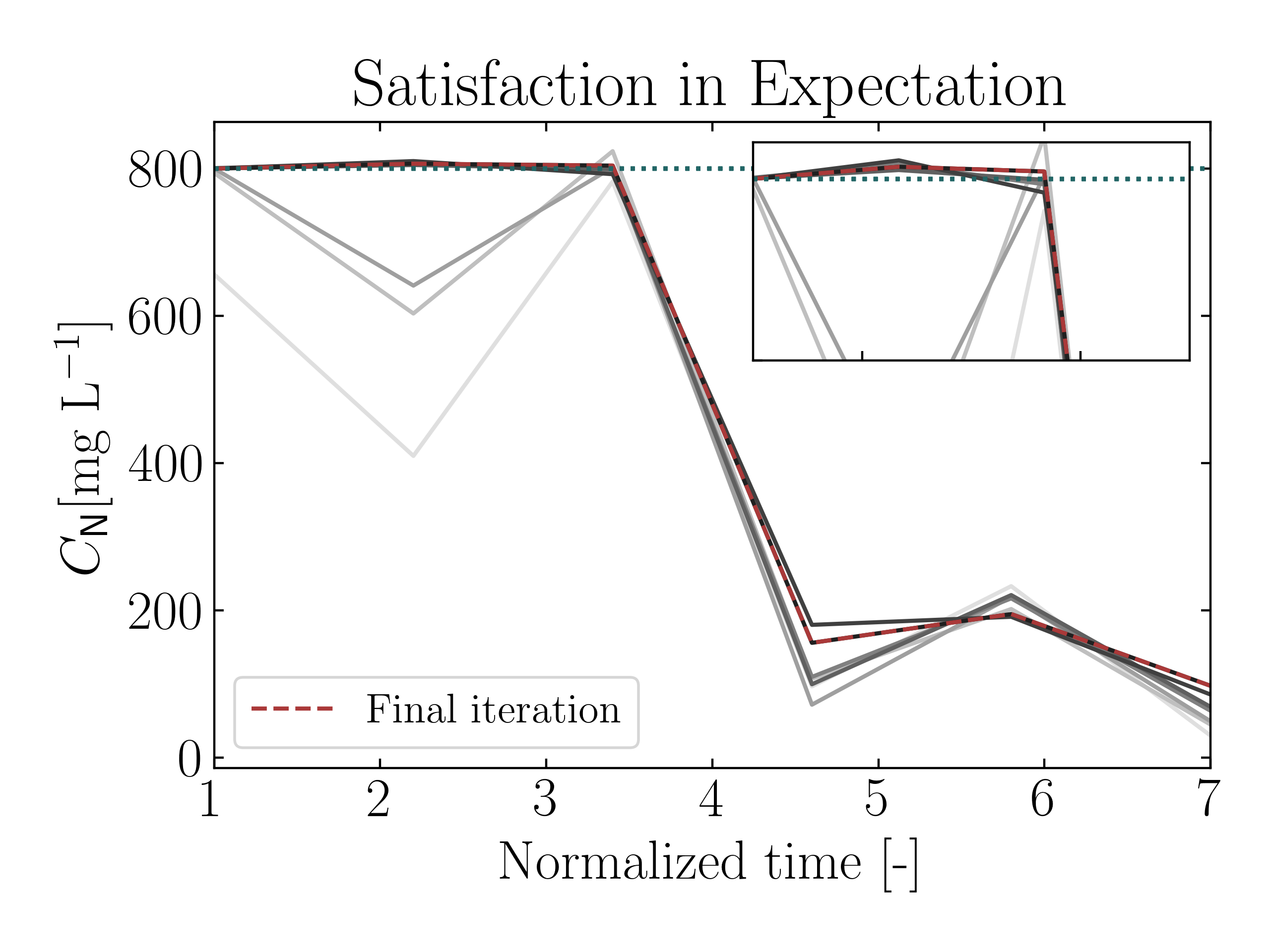}\vspace{-.85em}
  \caption{Constraint satisfaction in expectation  using trust-regions}
  \label{fig:expectation_tr}
\end{subfigure}
\begin{subfigure}{0.5\textwidth}
  \centering
  \includegraphics[width=.98\linewidth]{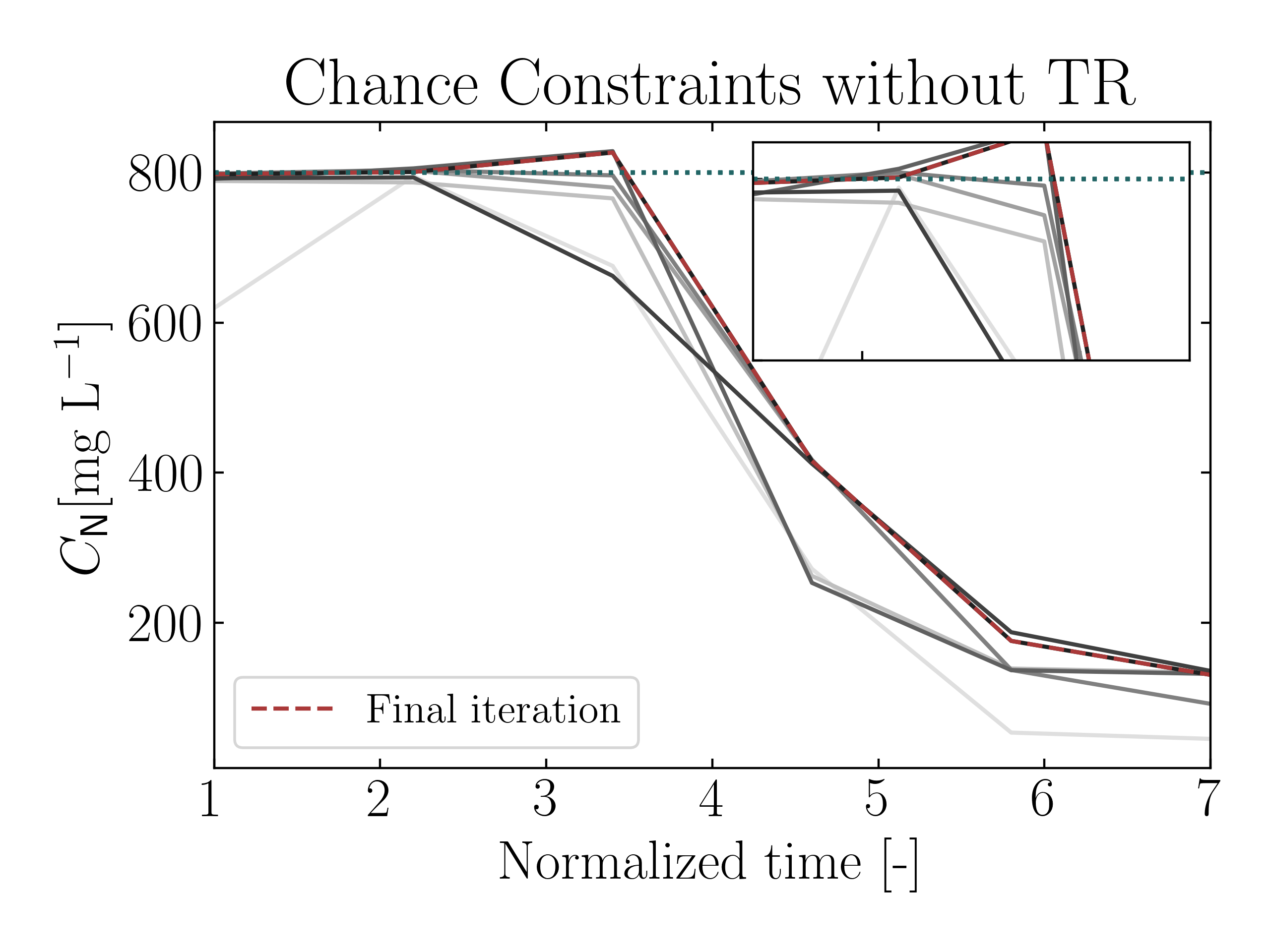}\vspace{-.85em}
  \caption{Chance constraint satisfaction without using trust-regions}
  \label{fig:PR_no_tr}
\end{subfigure}
\begin{subfigure}{0.5\textwidth}
  \centering
  \includegraphics[width=.98\linewidth]{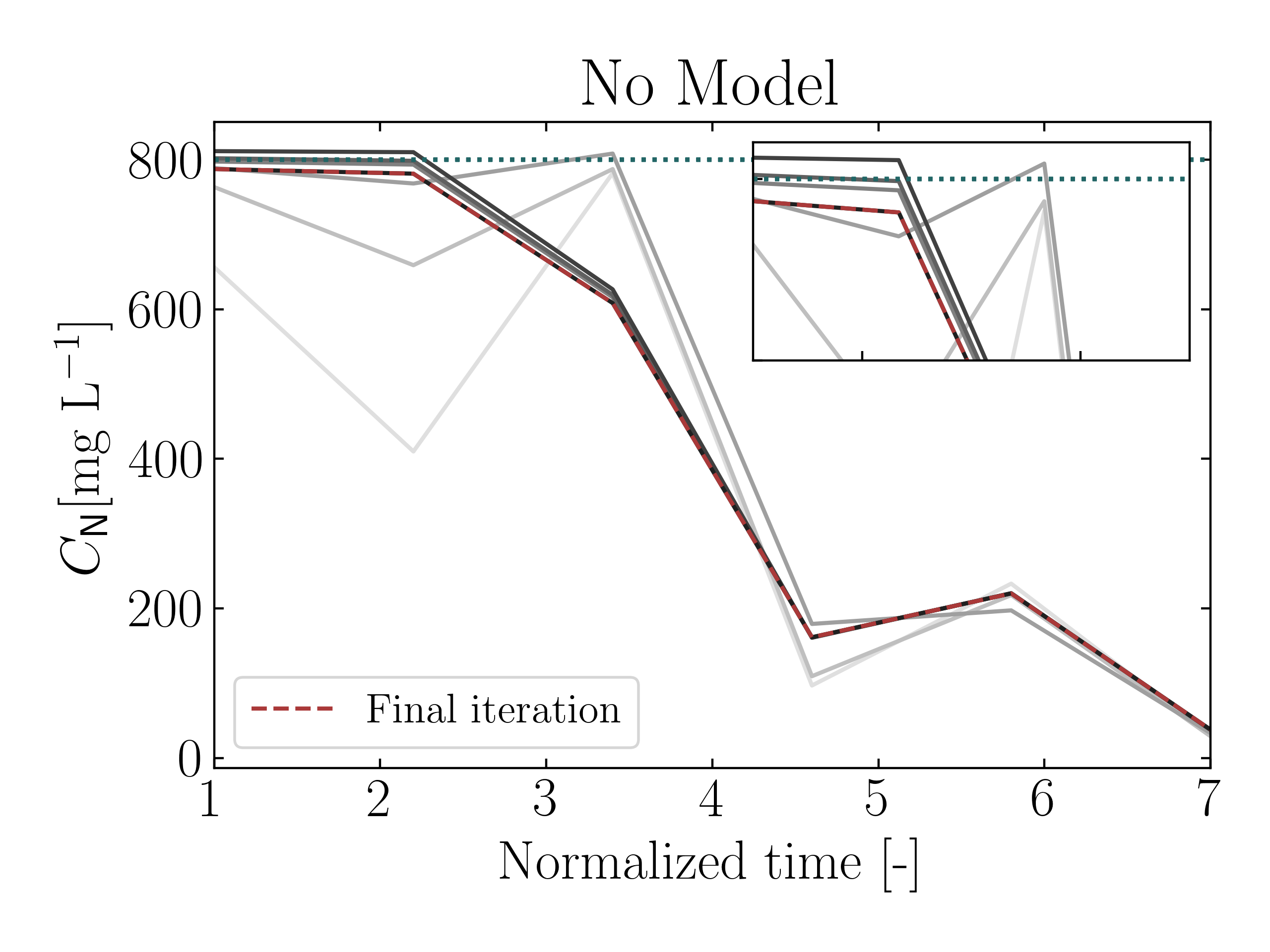}\vspace{-.85em}
  \caption{Chance constraint satisfaction without prior model using trust-regions}
  \label{fig:nomodel}
\end{subfigure}
\caption{RTO results for the photobioreactor case study (Problem~\ref{eq:PBR}) using two alternative approaches: {\bf (a)} The probabilistic constraints in (\ref{eq:modified_problem_GP+TR}) are replaced with the satisfaction in expectation. {\bf (b)} The trust-regions are removed from (\ref{eq:modified_problem_GP+TR}).{\bf (c)} The prior model is removed from (\ref{eq:modified_problem_GP+TR}).} 
\end{figure}
The constraint 1 ($C_N\leq 800$) for all the RTO iterations for a single MC of our proposed framework is depicted in Fig. \ref{fig:constraints},  where the lines are faded out towards earlier iterations. As expected the plant converges fast to a solution that satisfies the constraints and in the beginning the constraints are not close to the bound of the constraint.

The corresponding figures for the constraint using the expected value of the constraints (but with trust-regions), chance constraints without trust regions and the absence of prior model are depicted in Figure \ref{fig:expectation_tr} - \ref{fig:nomodel} respectively. 
Notice that the constraint violation occur often in all cases, where in Figure \ref{fig:PR_no_tr}, violates the constraints significantly for the case of the trust-region's absence. Such violation can explain the fast increase in the objective function, as the objective may obtain better values at infeasible regions.  

As depicted in  Fig. \ref{fig:constraints} and Fig. \ref{fig:expectation_tr} - \ref{fig:nomodel}, our proposed frameworks is superior in terms of constraints satisfaction.
Therefore it is clear that the proposed framework can significant beneficial real-time optimization, when stochastic environments are present and safe exploration is needed. 

\section{CONCLUSIONS}\label{sec:conc}

A new approach is proposed for real-time optimization under plant-model mismatch using multi-fidelity GPs and trust-regions to efficiently incorporate chance constraints even when only low fidelity models are available.
The low fidelity model assumed to be a black-box is emulated using a GP so that the optimization of the problem is solved more efficiently. 
The trust region restricts the design space which results in meaningful predictions for the mean and variance of the GPs, and allows us to reformulate the chance constraints with confidence.

The benefits are analyzed and illustrated with a numerical case study:
a challenging batch-to-batch optimization problem with a dozen inputs and constraints. Here we have shown that our proposed framework is a good trade-off between optimality and safe exploration. In practical applications, this added reliability could outweigh the benefits of model-free RTO, for instance, ease of design and maintainability. These results reflect that using a nominal model in the manner of a prior constitutes an effective de-risking strategy in higher-dimensional RTO problems.





\section*{ACKNOWLEDGMENT}
This project has received funding from the EPSRC grant project (EP/R032807/1) is gratefully acknowledged.
\bibliographystyle{IEEEtran}
\bibliography{IEEEabrv, IEEEexample.bib}


\end{document}